\documentclass{article}

\usepackage{microtype}
\usepackage{graphicx}
\usepackage{subfigure}
\usepackage{booktabs}
\usepackage{amsmath}
\usepackage{placeins}

\usepackage{hyperref}

\usepackage[accepted]{icml2021}

\icmltitlerunning{To Trust or Not to Trust a Regressor}

\begin{document}

\twocolumn[
\icmltitle{To Trust or Not to Trust a Regressor: Estimating and Explaining Trustworthiness of Regression Predictions}



\icmlsetsymbol{equal}{*}

\begin{icmlauthorlist}
\icmlauthor{Kim de Bie}{uva}
\icmlauthor{Ana Lucic}{uva}
\icmlauthor{Hinda Haned}{uva}
\end{icmlauthorlist}

\icmlaffiliation{uva}{University of Amsterdam, The Netherlands}

\icmlcorrespondingauthor{Kim de Bie}{kim.de.bie@outlook.com}

\icmlkeywords{Machine Learning, Explainability, ICML}

\vskip 0.28in
]



\printAffiliationsAndNotice{}  

\begin{abstract}
In hybrid human-AI systems, users often need to decide whether or not to trust an algorithmic prediction while the true error in the prediction is unknown.
To accommodate such settings, we introduce RETRO-VIZ, a method for (i) estimating and (ii) explaining trustworthiness of regression predictions.
It consists of RETRO, a quantitative estimate of the trustworthiness of a prediction, and VIZ, a visual explanation that helps users identify the reasons for the (lack of) trustworthiness of a prediction.
We find that RETRO-scores negatively correlate with prediction error. 
In a user study with $41$ participants, we confirm that RETRO-VIZ helps users identify \emph{whether} and \emph{why} a prediction is trustworthy or not.
\end{abstract}

\section{Introduction}
\label{introduction}

Machine learning algorithms are increasingly used in high-stakes domains.
This is not without risk: a trained model can produce erroneous predictions, especially when it is asked to generalize beyond situations it is familiar with \cite{amodei_concrete_2016}.
To prevent failures that would result from an AI system working autonomously, algorithms are often used in hybrid systems where humans aid in the decision-making process \cite{kamar_directions_2016}.

It remains unclear \emph{how} humans should recognize erroneous algorithmic predictions in production settings, where the true error in individual predictions is usually unknown.
Beyond global performance metrics, such as the error on a test set, many machine learning systems do not provide estimates of the reliability of individual predictions \cite{nushi_towards_2018}. 
While explainable AI (XAI) methods have been proposed as a tool for assessing the reliability of algorithmic predictions ~\cite{doshi-velez_towards_2017}, 
in practice, researchers have struggled to demonstrate that XAI methods help users do this \citep{lai_harnessing_2020}.
Overall, current methods are insufficient to help users understand when, and for what reason, algorithmic predictions are (not) trustworthy in production settings where the true error is unknown.

To address this gap, we propose \textbf{RE}gression \textbf{TR}ust sc\textbf{O}res with \textbf{VI}suali\textbf{Z}ations (RETRO-VIZ). It consists of two parts: (i) \textbf{RETRO}, a method for quantitative estimation of trustworthiness in regression predictions when the true error is unknown, and (ii) \textbf{VIZ}, a visualization that helps users identify the reasons for the estimated trustworthiness. 
The goal of RETRO-VIZ is to provide insight into algorithmic trustworthiness in a way that aids human-algorithm co-operation. 
This goal is based on the needs of a large international retailer headquartered in the Netherlands, where understanding the trustworthiness of individual algorithmic predictions 
is important for tasks such as forecasting sales or the effect of an upcoming promotion. 
Currently, relatively simple, transparent regression models are used for prediction tasks.
Although stakeholders have expressed interest in adopting more complex methods because of potential increases in overall performance, they have also expressed hesitation -- they are concerned that the lack of transparency in such methods makes it more difficult to assess how and when the model is making mistakes.

We aim to answer the following research questions:

\begin{itemize}
    
    \item[\textbf{RQ1}]
    Do the estimates of trustworthiness that \textbf{RETRO} produces correlate with the errors in algorithmic predictions, and if so, how?
    
    \item[\textbf{RQ2}]
    Under which conditions does \textbf{RETRO} perform best, given different (a) model architectures, (b) data dimensionalities and (c) causes of error?
    
    \item[\textbf{RQ3}]
    Do \textbf{VIZ}-explanations \textit{objectively} help users recognize whether and why algorithmic predictions are (un)trustworthy?
    
   \item[\textbf{RQ4}]
   Do users \textit{subjectively} evaluate \textbf{VIZ}-explanations as being valuable in practice?
    
\end{itemize}

\section{Related Work}

RETRO-VIZ bridges the gap between uncertainty estimation and explainability in machine learning: we provide an estimate of the trustworthiness of individual regression predictions, as well as an explanation for why a prediction is (not) trustworthy.
Here, we provide an overview of the existing work in the fields of uncertainty estimation and explainability, and explain why existing methods are insufficient to understand when and why algorithms fail in production settings, where the error is unknown.
In addition, we examine how trustworthiness is defined in different fields and relate this to the definition we use in this work.

\subsection{Uncertainty Estimation in Machine Learning}

Some algorithms, such as neural classifiers or Bayesian architectures, provide a measure of confidence in their predictions by default \citep{kendall_what_2017}.
When available, these uncertainty estimates can suffer from various limitations, such as poor calibration \cite{guo_calibration_2017} or unreliability \cite{goodfellow_explaining_2015}.
To address this, some approaches attempt to model uncertainty directly by adapting the architecture of the model \cite{gal_dropout_2016, papernot_deep_2018}.

In practice, it may not always be feasible to require a certain model architecture.
To this end, several model-agnostic methods have been proposed for estimating predictive uncertainty. 
\citet{jiang_trust_2018} proposed a `Trust Score', which estimates the trustworthiness of individual predictions made by a classifier.
This is done by measuring the agreement between a classifier and a modified nearest-neighbor classifier on a particular test instance. 
This concept is closely related to case-based reasoning (CBR) methods, which assume that similar problems should have similar solutions \citep{aamodt_case-based_1994, kenny2019twin, li2018deep}.

As \citet{rajendran_accurate_nodate} find, the Trust Score correlates with errors that are caused by distributional shift, but less with classification error arising from other causes.
Although \citet{rajendran_accurate_nodate} proposed an alternative method that captures a wider range of uncertainty types, their method suffers from the same limitations as \citet{jiang_trust_2018}: (i) both methods are only applicable to classification problems, and (ii) both methods do not provide any explanation as to \emph{why} a prediction is (not) trustworthy.
In contrast, RETRO-VIZ estimates the trustworthiness in regression predictions, and provides context for the estimated trustworthiness through a visual explanation.

\subsection{Explainable Machine Learning}

Increasing the interpretability of complex models can help users make better use of algorithmic predictions \citep{doshi-velez_towards_2017}.
Specifically, model explanations may help with debugging and detecting errors: if an algorithm is making decisions on the basis of irrelevant factors, this could be an indication that there is something wrong \citep{ribeiro_why_2016}.
In a study of how organizations use explanation methods in practice, \citet{bhatt_explainable_2019} found that detecting errors is the most common use case for explanations. 
However, researchers have struggled to demonstrate that explanations, which commonly focus on explaining model internals, lead to an improvement in decision quality \citep{lai_harnessing_2020}.

\citet{lucic2020does} developed a method for explaining errors produced by regression models, which is similar to what we propose.
However, this method can only be used to explain \emph{known} errors, i.e. errors for which the ground-truth target value is available, which is usually not the case when the model is first put in production. 
In contrast, RETRO-VIZ provides estimates as well as explanations of the trustworthiness of predictions when the true value is not (yet) known and can therefore be employed in production environments.

\citet{antoran2020getting} proposed CLUE, a method for creating counterfactual explanations of uncertainty estimates from Bayesian Neural Networks.
The counterfactuals provided indicate how an input would have to be changed such that the model becomes more certain about its prediction.
Unlike our method, CLUE relies on existing uncertainty estimates, and is constrained to Bayesian models alone. 
In contrast, RETRO-VIZ can provide explanations for any regressor. 

\subsection{Trustworthiness}
\label{definingtrust}
This work proposes a method that measures the \emph{trustworthiness} of regression predictions.
The concepts of trust and trustworthiness are used differently in various branches of machine learning scholarship.
As \citet{lipton_mythos_2017} points out, `improving trust' is an important motivation for the development of XAI methods and is featured prominently in existing work \cite{lundberg_unified_2017, ribeiro_why_2016}, but a clear definition of trust or trustworthiness is lacking in the XAI literature. 
The implicit understanding is that an algorithm is trustworthy if its predictions are based on factors that are acceptable to domain experts instead of on `spurious correlations' \citep{ribeiro_why_2016}.

In \citet{jiang_trust_2018}, a prediction is considered \emph{trustworthy} if it is reasonable in light of the training data, in the sense that the train data behaves in a similar way to the new instance.
As the method we propose for the numeric estimation of trustworthiness is heavily inspired by \citet{jiang_trust_2018}, we adopt this definition of trustworthiness.
We then explicitly distinguish between the \emph{trustworthiness} and the \emph{correctness} of a prediction:
trustworthiness expresses whether a prediction is aligned with the train data without having access to the ground truth, while correctness expresses whether a prediction is erroneous or not, which requires access to the ground truth.
However, we expect that trustworthy predictions tend to be more accurate because they are grounded in the training data, and we assess this in \textbf{RQ1} and \textbf{RQ2}.

\section{Method}
\label{retromethod}

In this section, we introduce RETRO-VIZ, a method for identifying and explaining the trustworthiness of regression predictions. 
It consists of two components: (i) RETRO, which provides a numerical estimation of trustworthiness for individual predictions, and (ii) VIZ, which visually explains the RETRO-score.
We build upon the method proposed by \citet{jiang_trust_2018}, which estimates trustworthiness in predictions by a classifier based on the idea that similar instances should receive a similar prediction.

First, we calculate the RETRO score. 
RETRO requires (i) a trained regression model, (ii) the data that the model was trained on $(X_{train}, Y_{train})$, and (iii) a new instance $x_p$ and its predicted target value $\hat{y}_p$.
All input and output variables must be numeric.
We treat the model as a black box and do not require access to the internals or parameters.
The method consists of three phases, which are discussed below.

\subsection{RETRO Phase 1: Preparing the Reference Set}

RETRO relies on a reference set which is based on the training data. 
As will be outlined in Phase 2, RETRO leverages the distance between the new instance and similar instances from this set to estimate the trustworthiness of a prediction.

\paragraph{Step 1A. Filter errors from the train data} 

If the regression model has been trained on data that is similar to the new instance, we assume that the model will be able to make a high-quality prediction for this instance.
However, this does not hold when the model has not sufficiently captured a region of the (training) input space.
Therefore, we remove those instances from the train set for which the model produces an (absolute) error in the top-$\alpha$ fraction of all training data, where $\alpha = 0.1$ in our experiments.

\paragraph{Step 1B. Reduce dimensionality of the data (optional)}
Since RETRO involves a k-Nearest Neighbors (kNN) procedure, which is known to perform poorly in high-dimensional spaces \cite{beyer_when_1999}, we reduce the dimensionality of the data when the number of features is greater than $15$. 
We do this by training a Multi-Layer Perceptron (MLP) whose penultimate layer represents the desired dimensionality ($10$ in our experiments). 
The MLP is trained to predict $Y_{train}$ given $X_{train}$.
In our experiments, we found that this method performs better than classic dimensionality-reduction methods such as PCAs or autoencoders. 
We postulate that this is because the MLP maintains those aspects of the original input most relevant to making the prediction.

\subsection{RETRO Phase 2: Relationship to Neighbors}

In Phase 2, we assess whether the predicted target value for the new instance seems trustworthy in light of the reference set created in Phase 1.

\paragraph{Step 2A. Find the closest neighbors of the new instance}
Using kNN, we find the $K$ instances in the reference set that are most similar to our new instance $x_p$ as measured through Euclidean distance.

\paragraph{Step 2B. Find the mean distance to the neighbors ($d_1$)}
For a trustworthy prediction, we expect that the new instance has neighboring instances at a relatively close distance in the reference set.
If this is not the case, the new instance is likely to be an outlier.
We measure this as $d_1$, which is the average Euclidean distance of the new instance to all of its neighbors.
The larger $d_1$, the less the new instance resembles the reference data, and therefore the lower the expected quality of the prediction.

\paragraph{Step 2C. Find the distance between the predicted and ground-truth targets ($d_2$)}
For a trustworthy prediction, we also expect that the target prediction for the new instance is close to the target value of the neighbors in the reference set.
We measure this as $d_2$, which is the distance of $\hat{y}_p$ to the mean ground-truth value of the neighbors.
     \begin{equation}
         d_2 = \left| \left( \dfrac{1}{K} \sum_{k=1}^K y_k \right) - \hat{y}_p \right|
     \end{equation}{}
As we expect similar instances to receive similar predictions, $d_2$ should be small for high-quality predictions.

\subsection{RETRO Phase 3: Scoring and Normalization}

In Phase 3, we determine the RETRO-score for a prediction and optionally normalize it.

\paragraph{Step 3A. Calculate the RETRO-score $R$ from $d_1$ and $d_2$}
The RETRO-score $R$ is defined as follows:
\begin{equation}
    R = - \left(\beta \cdot d_1 + \left[1 - \beta\right] \cdot d_2 \right)
\end{equation}{}
Here, $\beta$ may be used to weight the contribution from $d_1$ relative to $d_2$. Based on our experiments, we found $\beta = 0.5$ to be a good default value.

\paragraph{Step 3B. Normalize the RETRO-score (optional)}
$R$ as it is calculated in Step 3A has a potential range of $\left[-\infty, 0\right]$, given that both $d_1$ and $d_2$ are unbounded.
This makes $R$ difficult to interpret in isolation since its value is largely dependent on the particular dataset and model.
To normalize $R$ to a range of  $\left[0, 1\right]$, we calculate the RETRO-score $\Tilde{R}$ for each of the instances in the training set and obtain the minimum and the maximum score, $\Tilde{R}_{min}$ and $\Tilde{R}_{max}$. 
We normalize the RETRO-scores for new instances to this range, where values that fall outside the range are clipped to $0$ or $1$.
Note that this normalization is optional, and the need to do so is dependent on the domain.

\begin{figure*}[ht]
\begin{center}
\centerline{\includegraphics[width=.75\textwidth]{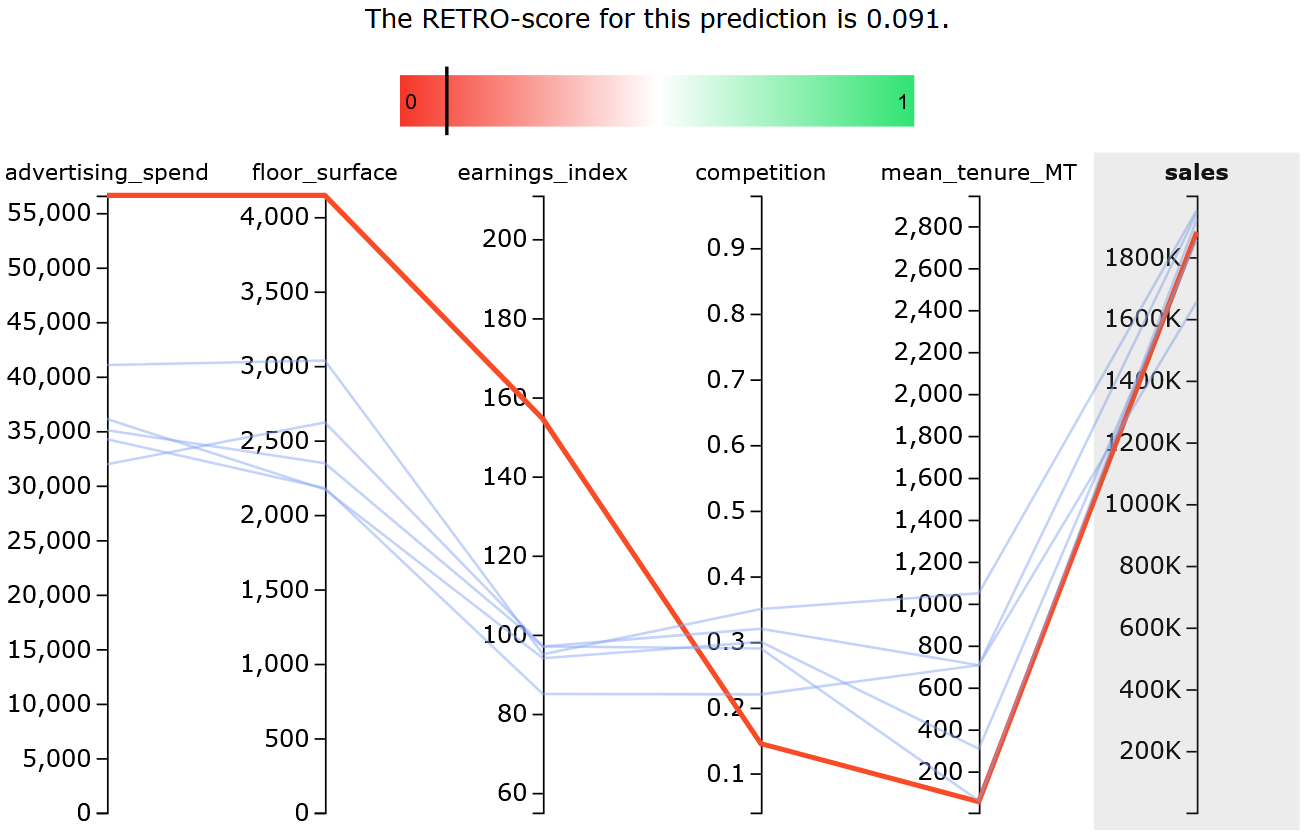}}
\caption{Example RETRO-VIZ output for an untrustworthy prediction. This model predicts sales based on five features. The RETRO-score is 0.091 for the test instance (red). The VIZ-explanation shows that its nearest neighbors from the reference set (blue) have feature values that are quite different from its own, even though the predicted sales value for the new store is similar to the ground-truth sales values of the neighboring instances. This indicates that the new instance is not well-represented in the training set.}
\label{vizplot}
\end{center}
\end{figure*}

\subsection{VIZ: Visually Explaining Trustworthiness}

\begin{figure*}[p]
\begin{center}
\centerline{\includegraphics[width=.75\textwidth]{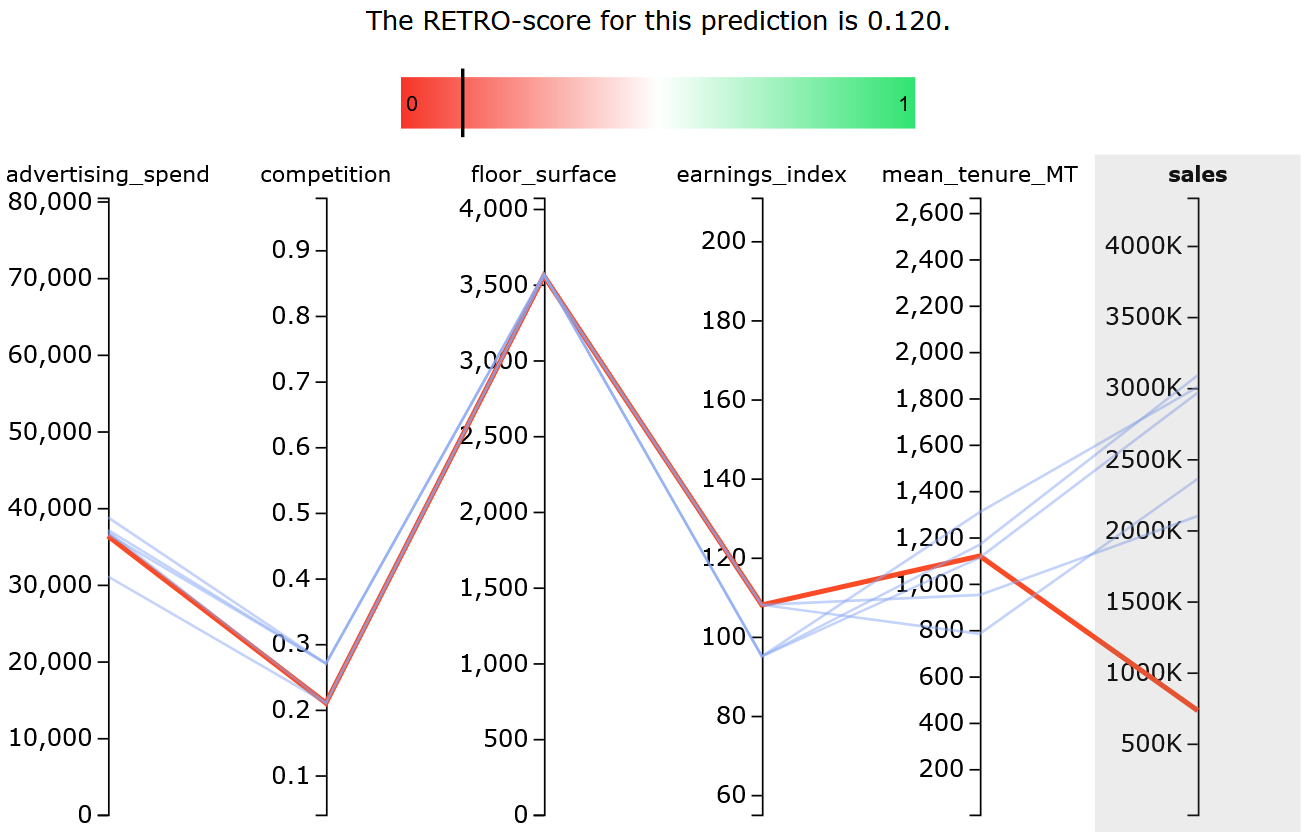}}
\caption{Example RETRO-VIZ output for an untrustworthy prediction. The model used is identical to that described in Figure \ref{vizplot}. The RETRO-score is 0.120 for the test instance (red). The VIZ-explanation shows that its nearest neighbors from the reference set (blue) have similar feature values, but the predicted sales for the new instance are much larger. This indicates that the prediction is untrustworthy.}
\label{vizplot-lowpred}
\end{center}
\end{figure*}

\begin{figure*}[p]
\begin{center}
\centerline{\includegraphics[width=.75\textwidth]{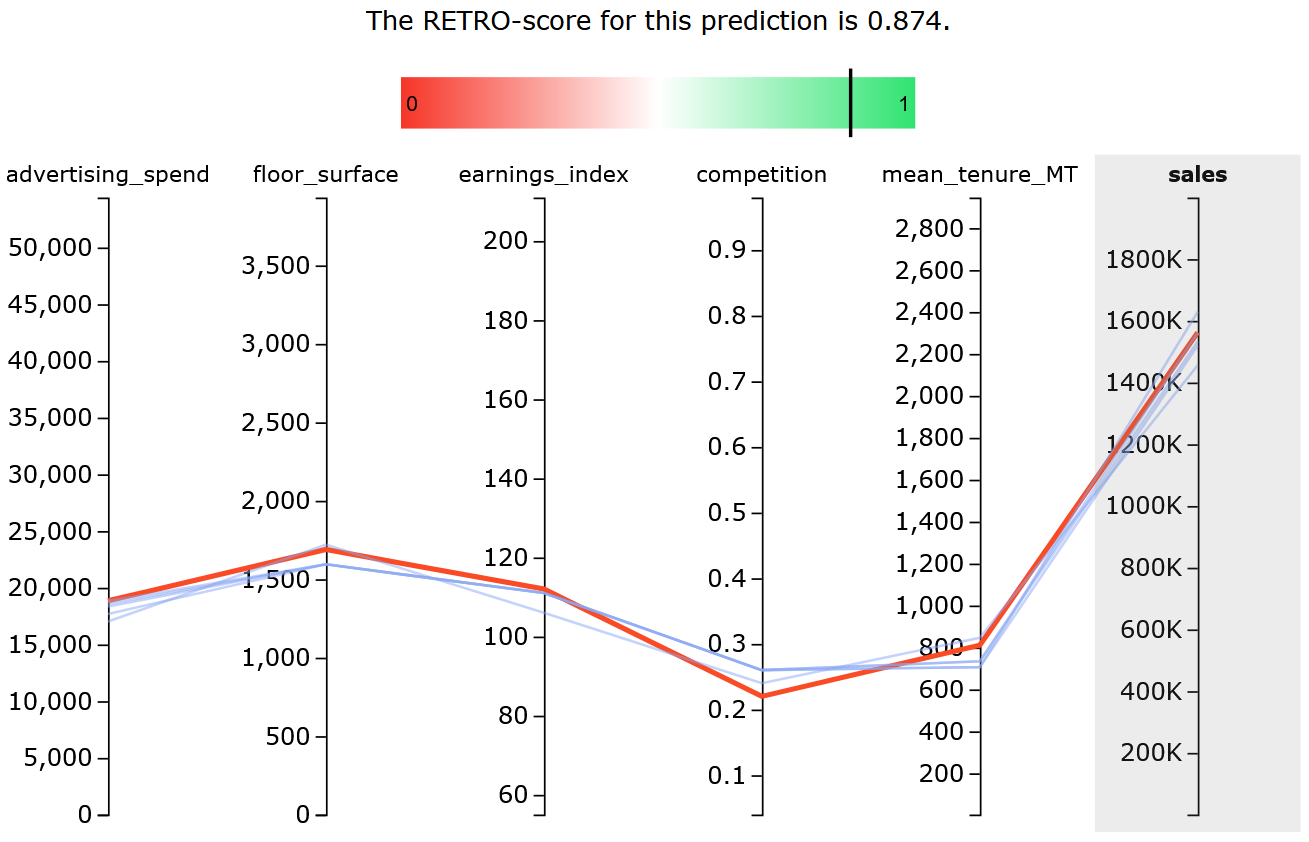}}
\caption{Example RETRO-VIZ output for a trustworthy prediction. The model used is identical to that described in Figure \ref{vizplot}. The RETRO-score is 0.874 for the test instance (red). The new instance is well-aligned with its neighbors from the test set and therefore the prediction seems trustworthy.}
\label{vizplot-high}
\end{center}
\end{figure*}

Besides identifying predictions that are potentially erroneous, we want to provide users with an actionable tool that helps them understand \textit{why} a particular prediction is (un)trustworthy. 
We use Parallel Coordinate Plots, as proposed by \citet{inselberg_plane_1985} for displaying multi-dimensional data, to visualize the new instance and its $K$ neighbors as retrieved through the RETRO-method.
Example VIZ-plots are provided in Figures \ref{vizplot}, \ref{vizplot-lowpred} and \ref{vizplot-high}.
Note that in practice, these plots are interactive: users can observe the exact feature values of each instance (represented by a line) by hovering the cursor over the line. 
While dimensionalities of up to $10$ can be reasonably displayed on a single VIZ-plot, features have to be selected for higher-dimensional feature spaces, for example by selecting important features though LIME \cite{ribeiro_why_2016} or SHAP \cite{lundberg_unified_2017}, or based on user expertise.

VIZ-plots make it straightforward to identify why a prediction has received a high or low RETRO-score.
A low RETRO-score can be obtained for a new instance for two main reasons.
Firstly, it could be that the new instance has no instances in the reference set that lie relatively close to it, because the new instance deviates substantially in one or multiple features.
An example of this is provided in Figure \ref{vizplot}.
Secondly, it could be that the predicted target variable lies far away from the ground-truth target variables of the $K$ neighbors (see Figure \ref{vizplot-lowpred} for an example).
In contrast, when a new instance and its prediction are aligned with the reference data, they will receive a high RETRO-score, and the lines on the VIZ-plot will lie close to each other, like in Figure \ref{vizplot-high}.

VIZ-plots allow users to leverage their domain expertise: based on the way in which the new instance deviates from its neighbors, the domain expert can design an appropriate response.
For example, in Figure \ref{vizplot}, we observe that the new instance is a much larger store (measured by floor\_surface), and spends more on advertising than the most similar stores in the reference set, so it is likely that true sales are higher than predicted by the model.

\section{Evaluating the RETRO-score}
\label{retroeval}

\begin{table*}[!ht]
\centering
\small
\begin{tabular}{@{}lrrrrrrrrrr@{}}
\toprule
                       & \multicolumn{5}{c}{\textit{(i) Distributional shift}}                                                                                                                         & \multicolumn{2}{c}{\textit{(ii) Model Overfit}}                               & \multicolumn{2}{c}{\textit{(iii) Model Underfit}}                                & \multicolumn{1}{l}{}              \\ \cmidrule(lr){2-6}
\cmidrule(lr){7-8}
\cmidrule(lr){9-10}
\textbf{Dataset (features)}        & \multicolumn{1}{c}{\textbf{LR}} & \multicolumn{1}{c}{\textbf{SVR}} & \multicolumn{1}{c}{\textbf{MLP}} & \multicolumn{1}{c}{\textbf{DT-15}} & \multicolumn{1}{c}{\textbf{RF}} & \multicolumn{1}{c}{\textbf{GP}} & \multicolumn{1}{c}{\textbf{DT-10k}} & \multicolumn{1}{c}{\textbf{MLP-S}} & \multicolumn{1}{c}{\textbf{DT-1}} & \multicolumn{1}{c}{\textbf{Mean (SD)}} \\ \midrule
\textit{Cyclepower} (4)                              & -0.947                          & -0.370                           & -0.690                           & -0.327                             & -0.372                          & $-0.282^\dagger$                          & -0.389                              & -0.906                              & -0.741                            & \textbf{-0.593 (0.258)}                   \\
\textit{Airfoil} (5)                                     & -0.880                          & -0.739                           & -0.697                           & -0.495                             & -0.567                          & -0.999                          & -0.199                              & -0.909                              & -0.393                            & \textbf{-0.653 (0.263)}                   \\
\textit{Store sales**} (5)                                     & -0.062                          & -0.782                           & -0.663                           & -0.614                             & -0.650                          & -1.000                          & -0.534                              & -0.780                              & -0.834                            & \textbf{-0.658 (0.262)}                   \\
\textit{Fish toxicity}  (6)                                     & -0.695                          & -0.589                           & -0.563                           & -0.460                             & -0.419                          & -1.000                          & -0.414                              & -0.791                              & -0.456                            & \textbf{-0.599 (0.198)}                   \\
\textit{Abalone}   (7)                                     & -0.634                          & -0.427                           & -0.399                           & -0.063                             & -0.109                          & -0.999                          & -0.609                              & -0.552                              & -0.212                            & \textbf{-0.445 (0.295)}                   \\
\textit{Autompg}  (7)                                     & -0.566                          & -0.892                           & -0.954                           & -0.017                             & -0.210                          & -0.992                          & -0.542                              & -0.914                              & -0.586                            & \textbf{-0.630 (0.345)}                   \\
\textit{Cal. housing*}  (8)                                     & -0.995                          & -0.607                           & -0.964                           & -0.263                             & -0.261                          & -1.000                          & -0.475                              & -0.271                              & -0.433                            & \textbf{-0.585 (0.322)}                   \\
\textit{Energy eff.}    (8)                                     & -0.980                          & -0.415                           & -0.948                           & -0.472                             & -0.409                          & -0.734                          & -0.225                              & -0.940                              & -0.458                            & \textbf{-0.620 (0.284)}                   \\
\textit{Diabetes*}   (10)                                & -0.056                          & -0.237                           & -0.246                           & -0.341                             & -0.305                          & -0.751                          & -0.255                              & -0.460                              & -0.147                            & \textbf{-0.311 (0.200)}                   \\
\textit{Wine quality}   (11)                                    & -0.163                          & -0.289                           & -0.512                           & -0.483                             & -0.234                          & -0.999                          & -0.448                              & -0.311                              & -0.212                            & \textbf{-0.406 (0.255)}                   \\
\textit{Boston*} (13)                                    & -0.348                          & -0.863                           & -0.793                           & -0.864                             & -0.657                          & -0.699                          & -0.433                              & -0.687                              & -0.412                            & \textbf{-0.640 (0.197)}                   \\
\textit{Supercond.} (81)                    & -0.946                          & -0.275                           & -0.361                           & -0.546                             & -0.511                          & -1.000                          & -0.463                              & -0.719                              & -0.745                            & \textbf{-0.618 (0.251)}                   \\
\textit{Communities} (100)                              & -0.880                          & -0.323                           & -0.186                           & -0.718                             & -0.408                          & -0.617                          & -0.604                              & -0.620                              & -0.535                            & \textbf{-0.543 (0.210)}                   \\  \cmidrule(lr){2-6}
 \cmidrule(lr){7-8}
 \cmidrule(lr){9-10}
  \cmidrule(lr){11-11}
\textit{\textbf{Mean (SD)}}                      & \textbf{-0.627}                 & \textbf{-0.524}                  & \textbf{-0.614}                  & \textbf{-0.436}                    & \textbf{-0.393}                 & \textbf{-0.899}                 & \textbf{-0.430}                      & \textbf{-0.682}                     & \textbf{-0.474}                   & \textit{\textbf{-0.561}}          \\
 & \textbf{(0.358)}                 & \textbf{(0.234)}                  & \textbf{(0.264)}                  & \textbf{(0.238)}                    & \textbf{(0.170)}                 & \textbf{(0.150)}                 & \textbf{(0.134)}                      & \textbf{(0.228)}                     & \textbf{(0.214)}                   & \textit{\textbf{(0.268)}}          \\ \bottomrule
\end{tabular}
\caption{Pearson correlation coefficient $\rho$ between the absolute error and the RETRO-score for all experimental settings. Correlations closer to $-1$ are more desirable. The number of features is provided in the first column. See Section \ref{retroeval} for the model acronyms. *: Datasets from \textit{scikit-learn}. **: Internal company dataset. Other datasets from UCI-ML \cite{dua_uci_2020}. $\dagger$ GP did not lead to overfitting, so the value was not included in the mean and standard deviation.}
\label{tab:allresults}
\end{table*}

\subsection{Experimental Setup}

To answer \textbf{RQ1}, we assess the Pearson correlation coefficient $\rho$ between the RETRO-score and the absolute error in predictions (which is unknown in production settings).
We expect a negative correlation: a lower RETRO-score, which implies a lower trustworthiness of predictions, should correspond to larger errors.
To answer \textbf{RQ2}, we assess the correlation across a wide range of models (\textbf{RQ2a}), data dimensionalities (\textbf{RQ2b}), and error causes (\textbf{RQ2c}).
Following \citet{rajendran_accurate_nodate}, we select models and datasets to evaluate specific causes of error: (i) \textit{Distributional Shift}, (ii) \textit{Model Overfit} and (iii) \textit{Model Underfit}, resulting in $117$ experimental settings. 
We use these error causes because they represent common sources of algorithmic failure, both during model development (overfit/underfit) and when the model is in production (distributional shift).
We use 13 datasets in our experiments, which originate from \textit{scikit-learn}\footnote{\url{https://scikit-learn.org/stable/}}, the UCI ML repository \citep{dua_uci_2020} and our company.
We use $80\%$ of the data to train the models and $20\%$ as test data on which we evaluate RETRO.

\paragraph{Error Cause (i): Distributional Shift}
Distributional shift occurs when a model produces erroneous predictions on a test set because it was trained on data from a different distribution. 
Three of our datasets, \textit{Store sales}, \textit{Wine quality} and \textit{Autompg}, can be split in such a way that distributional shift naturally occurs.
For example, the \textit{Wine quality} dataset consists of white and red wines, so we train on white and test on red wines.

For the remaining datasets, we induce distributional shift manually.
To do this, we select the $30\%$ of most important features in the test set according to their SHAP value \cite{lundberg_unified_2017}.
Then, for each feature $k$, we find its maximum value, $x_k^{max}$, and generate noise for each instance $n_k^i$ from a Gaussian distribution, where $n_k^i \sim \mathcal{N}(x_k^{max},\,0.1 \cdot x_k^{max})$. 
We then add this noise to the original feature, so that $\tilde{x}_k^i = x_k^i + n_k^i$.
In this way, we ensure that the values of the perturbed features lie outside the original distribution.

We train five types of models: a Linear Regression (LR), a Support-Vector Regressor (SVR), a Multi-Layer Perceptron (MLP) with layer sizes $\left[50, 20, 10\right]$, a Decision Tree (DT), and a Random Forest (RF) with 100 trees. Both the DT and RF a have maximum depth of $15$.

\paragraph{Error Cause (ii): Model Overfit}
Model overfit occurs when a model fits the train data `too well' and picks up on randomness that is not part of the underlying distribution. 
To evaluate the RETRO-score for predictions suffering from this type of error, we train a very deep DT (depth 10,000; DT-10k) and a Gaussian Process (GP) Regressor with an RBF kernel in such a way that they produce a near-zero error on the train data, but a much larger error on test data.

\paragraph{Error Cause (iii): Model Underfit}
When a model is not complex enough to capture the patterns present in the data, it \emph{underfits} the data and produces poor-quality predictions on both the train and test set.
We induce this error by fitting a very shallow MLP with one hidden layer of width 10 trained for 3 iterations (MLP-S) and a DT of depth 1 (DT-1) to the train data.

\subsection{Results}

An overview of all $117$ experimental settings is given in Table \ref{tab:allresults}, which lists the Pearson correlation coefficient $\rho$ for each dataset and model.
In all cases, $\rho$ is negative, which confirms our expectations: RETRO-scores are lower when the prediction error is larger.
Across all settings, the average value of $\rho$ is $\mathbf{-0.561}$.
Thus, similar to \citet{jiang_trust_2018}, we confirm that trustworthiness as measured by RETRO correlates with correctness as measured by the absolute error in predictions.
While this correlation is not perfect, it shows that RETRO is helpful as an indicator to help users distinguish erroneous predictions, which answers \textbf{RQ1}.

To answer \textbf{RQ2}, we study the conditions under which RETRO performs best by examining the magnitude of $\rho$ across settings. 

\subsubsection{RQ2a: Performance across model architectures}
We first examine for which model architectures RETRO performs best. 
We obtain the best performance for predictions generated by an overfitting GP (mean of $\mathbf{-0.899}$), the second-best performance for the underfit MLP-S ($\mathbf{-0.682}$), followed by the non-tree distributional shift models (mean of $\mathbf{-0.627}$ for LR, $\mathbf{-0.524}$ for SVR and $\mathbf{-0.614}$ for MLP).

We find that RETRO performs worse for tree-based models than for other model architectures.
This is to be expected, because tree-based models
exclusively predict from a set of values that is based on the training data.
As a result, component $d_2$ of the RETRO-score is always reasonably small.
Therefore, for tree-based models, the RETRO-score leans heavily on component $d_1$, which measures the distance between the nearest neighbors and the new instance.
We contrast this to other types of model architectures, whose predictions are not constrained in this way and can take more extreme values, leading to potentially larger values of $d_2$. 

\subsubsection{RQ2b/c: Performance across data dimensionalities and causes of error}
Next, we examine the effect of dimensionality on the RETRO-score.
In our experiments, the number of independent features in the datasets used ranges from 4 to 100. 
In our settings, we find that the number of features does not have a clear impact on the performance of the method.
This can be observed in the right-most column of Table \ref{tab:allresults}: we do not see a clear trend across the feature sizes.
The same holds for the three causes of error.
As noted earlier in this section, the three models for which RETRO performs best (GP, MLP-S and LR) span the three different error causes used in our experiments.
Likewise, for the tree-based models, we find an approximately equal performance across all error causes.

\begin{table}[]
\centering
\small
\begin{tabular}{@{}lccc@{}}
\toprule
\multicolumn{2}{l}{\textbf{}}                    & \textbf{Correct} & \textbf{$\chi^2$} \\ \midrule
                         & \textit{\textbf{SB1}} & 95.1\%               & 33.390**     \\
\textit{Select Best}     & \textit{\textbf{SB2}} & 97.6\%           & 37.098**     \\
                         & \textit{\textbf{SB3}} & 92.7\%               &  29.878**     \\ \cmidrule{1-4}
                         & \textit{\textbf{IR1}} & 68.3\%               &  5.488*       \\
\textit{Identify Reason} & \textit{\textbf{IR2}} & 82.9\%                       & 26.561**     \\
                         & \textit{\textbf{IR3}} & 75.6\%               &  10.756**     \\ \bottomrule
\end{tabular}
\caption{Correct answers to the objective questions. Significant differences denoted using * ($p \leq 0.05$) and ** ($p \leq 0.01$).}
\label{tab:userst_resultsquestions}
\end{table}

\section{Evaluating VIZ-explanations}
\label{vizeval}
In practice, it is users who decide (not) to use an algorithm or an explanation method. 
Indeed, ``if the users do not trust a model [...], they will not use it'' \citep{ribeiro_why_2016}.
For this reason, we are interested in objectively as well as subjectively evaluating RETRO-VIZ with users. Specifically, we want to understand (i) whether VIZ helps users to recognize the trustworthiness of predictions (\textbf{RQ3}), and (ii) whether VIZ is subjectively perceived as satisfying by potential users (\textbf{RQ4}).
We evaluate VIZ-plots in a user study with $41$ participants, all of whom are in-company data scientists or analysts.
To emulate a realistic and familiar setting, we generate VIZ explanations for a model trained on an internal store sales dataset.
We use a \emph{human-grounded} evaluation \citep{doshi-velez_towards_2017}: a simplified task to assess whether VIZ-explanations allow users to recognize untrustworthy predictions.
Here, the task is to predict store sales based on five input variables (see Figure \ref{vizplot}).

\begin{table*}[t]
\centering
\small
\begin{tabular}{@{}lccc@{}}
\toprule
  & \textbf{Agree} & \textbf{Disagree} & \textbf{$\chi^2$} \\ \midrule
Q1. The visualization shows me how accurate the prediction is. & 63.4\%                & 29.3\%                & 5.16*         \\
Q2. The visualization lets me judge when I should trust or not trust the prediction. & 87.8\%                & 7.3\%                 & 27.92**     \\
Q3. From the visualization, I understand why the algorithmic prediction is trustworthy. & 78.0\%                & 19.5\%                 & 14.40**     \\
Q4. The visualization is satisfying: I understand what it is showing. & 87.8\%                & 4.9\%                 & 30.42**     \\
Q5. The visualization gives me all information I need to assess the trustworthiness of a prediction. & 34.1\%                & 43.9\%                & 0.50       \\
Q6. I think the visualization of whether the prediction should be trusted is useful in operational settings. & 87.8\%                & 4.9\%                 & 30.42**     \\\bottomrule
\end{tabular}
\caption{Percentage of users who (strongly) Agree or (strongly) Disagree with each statement. As neutral answers are omitted, the total is not $100\%$. Significant differences are denoted using * ($p \leq 0.05$) and ** ($p \leq 0.01$).}
\label{ESSquestions}
\end{table*}

\subsection{Objective evaluation}
In the first part, we introduce the method and evaluate whether VIZ-plots are \emph{objectively} understood by participants.
We do this by showing users a pair of plots with differing RETRO-scores and asking them to indicate which plot corresponds to a more trustworthy prediction.
We \emph{do not} show the RETRO-score to the users because this may bias users to simply choose the plot with the lower RETRO-score -- we only use the RETRO-score to determine which prediction is more trustworthy. We repeat with three pairs of plots. 
We refer to these questions as \textit{Select Best (SB)}.

Next, we show users a single VIZ-plot and ask them to explain in words \emph{why} they believe the prediction to be (un)trustworthy.
We refer to these questions as \emph{Identify Reason (IR)}.
We do this with three different plots (IR1-3). 
In IR1, we show a sales prediction which strongly deviates from the ground-truth sales for the most similar instances in the train set (similar to Figure \ref{vizplot-lowpred}), so we expect users to refer to this deviation in the target variable.
In IR2 we show a prediction similar to that in Figure \ref{vizplot}, and we expect users to refer to the fact that the new instance does not resemble the train data.
In IR3, we show a prediction similar to Figure \ref{vizplot-high}, and expect users to argue the prediction is trustworthy because it is aligned with the train data.

\subsection{Subjective evaluation}
In the second part of the user study, we use an adapted version of the Explanation Satisfaction Scale as proposed by \citet{hoffman_metrics_nodate} to evaluate participants' \emph{subjective} attitudes towards the VIZ-plots. 
This scale attempts to standardize existing approaches to subjective evaluation of XAI methods, and measures ``the degree to which users feel that they understand [...] the process being explained to them'' \citep{hoffman_metrics_nodate}.
We slightly adapt the questions from the Explanation Satisfaction Scale to fit our context better. 
The adapted questions are listed in Table \ref{ESSquestions}.
Answers are provided on a five-point Likert scale.

\subsection{Results}

In this section, we discuss the results of the objective and subjective evaluation as performed in the user study.

\subsubsection{Results of objective evaluation}
Table \ref{tab:userst_resultsquestions} summarizes participants' answers to the objective questions. 
We find that for all objective questions, the vast majority of users are able to answer the question correctly, for both \textit{Select Best} and \textit{Identify Reason}.
Using a $\chi^2$ test, we find that this majority is significant for all questions ($p \leq 0.05$).
This answers \textbf{RQ3}: VIZ-plots \emph{objectively} help users understand whether \textit{(Select Best)} and why \textit{(Identify Reason)} algorithmic predictions are trustworthy.
Then, perhaps surprisingly, it seems that VIZ-plots alone are sufficient for users to distinguish trustworthy predictions from untrustworthy ones, even when users do not have access to the associated RETRO-score.

\subsubsection{Results of subjective evaluation}
In Table \ref{ESSquestions}, we show the subjective questions from the Explanation Satisfaction Scale, adapted to fit our context from \citet{hoffman_metrics_nodate}, and summarize the responses.
We omit neutral responses. 
For 5 out of 6 questions, a significantly larger number of participants agree with the statement rather than disagree ($p \leq 0.05$).
We find that participants in the user study experience VIZ-plots as understandable (\textbf{Q4}) tools to discover whether (\textbf{Q2}) and why (\textbf{Q3}) algorithmic predictions are trustworthy and accurate (\textbf{Q1}), and that they believe VIZ-plots could be valuable in their work (\textbf{Q6}; $p \leq 0.05$ for all questions).
There is no significant difference in the proportion of participants agreeing and disagreeing with \textbf{Q5}, indicating that further tools and analysis about whether to trust a prediction may be required.
This is not unexpected: to determine whether deviations in either features or predictions are problematic and to decide what actions should be taken, users must rely on their domain expertise and may leverage other sources of information within their organization. 
This answers \textbf{RQ4}: 
participants clearly \emph{subjectively} experience VIZ-plots as valuable to deal with the uncertainty in algorithmic predictions.

\section{Discussion and Conclusion}
We propose RETRO-VIZ, a method for assessing the trustworthiness of regression predictions.
In all $117$ experimental settings, we found a negative correlation between the RETRO-score for a prediction and its correctness. 
This shows that RETRO can help to gauge the trustworthiness of a prediction in the absence of the ground-truth error. 
In addition, in our user study, we find that VIZ-plots help users distinguish whether and why algorithmic predictions are trustworthy.
However, similarly to other XAI methods \cite{lakkaraju_how_2019, slack2020fooling}, RETRO-VIZ must be used with caution to avoid misguided trust.
While RETRO-scores generally correlate with error, the correlation is not perfect, and therefore trustworthiness as measured by RETRO does not equal correctness.
Therefore, user awareness of this limitation is crucial.

In future work, we aim to explore ways to make RETRO-VIZ applicable to settings beyond regression with quantitative features, such as regression with categorical features and time series regression problems.
In addition, we aim to expand on our user study by investigating to what extent RETRO-VIZ leads to a better end-to-end system performance as compared to other explainability methods. 
Lastly, building on our finding that users seem to be able to identify untrustworthy predictions from a VIZ-plot alone, without having access to the RETRO-score (see Section \ref{vizeval}), we aim to explore how RETRO and VIZ can be used in a complementary way in deployed hybrid human-AI systems.
For example, RETRO-scores could be used as an automated monitoring tool, where only for predictions whose RETRO-score is below a threshold a VIZ-plot is shown to the user. 

\FloatBarrier


\FloatBarrier
\section*{Open-Source Software and Data}
RETRO-VIZ can be installed via pip as the Python package `retroviz'.
Code for the experiments is also available at ~\url{https://github.com/kimdebie/retroviz}.

\section*{Acknowledgments}
This research was supported by the Netherlands Organisation for Scientific Research (NWO) under project nr. 652.001.003. All content represents the opinion of the authors, which is not necessarily shared or endorsed by their respective employers and/or sponsors. 
Part of this research was conducted while Kim de Bie and Hinda Haned were employed at Ahold Delhaize.
We thank all users for their participation in the user study.

\bibliography{references}

\begin{thebibliography}{27}
\providecommand{\natexlab}[1]{#1}
\providecommand{\url}[1]{\texttt{#1}}
\expandafter\ifx\csname urlstyle\endcsname\relax
  \providecommand{\doi}[1]{doi: #1}\else
  \providecommand{\doi}{doi: \begingroup \urlstyle{rm}\Url}\fi

\bibitem[Aamodt \& Plaza(1994)Aamodt and Plaza]{aamodt_case-based_1994}
Aamodt, A. and Plaza, E.
\newblock Case-based reasoning: Foundational issues, methodological variations,
  and system approaches.
\newblock \emph{AI Communications}, 7\penalty0 (1):\penalty0 39--59, 1994.

\bibitem[Amodei et~al.(2016)Amodei, Olah, Steinhardt, Christiano, Schulman, and
  Man{\'e}]{amodei_concrete_2016}
Amodei, D., Olah, C., Steinhardt, J., Christiano, P., Schulman, J., and
  Man{\'e}, D.
\newblock Concrete problems in ai safety.
\newblock \emph{arXiv preprint arXiv:1606.06565}, 2016.

\bibitem[Antor{\'a}n et~al.(2020)Antor{\'a}n, Bhatt, Adel, Weller, and
  Hern{\'a}ndez-Lobato]{antoran2020getting}
Antor{\'a}n, J., Bhatt, U., Adel, T., Weller, A., and Hern{\'a}ndez-Lobato,
  J.~M.
\newblock Getting a clue: A method for explaining uncertainty estimates.
\newblock \emph{arXiv preprint arXiv:2006.06848}, 2020.

\bibitem[Beyer et~al.(1999)Beyer, Goldstein, Ramakrishnan, and
  Shaft]{beyer_when_1999}
Beyer, K., Goldstein, J., Ramakrishnan, R., and Shaft, U.
\newblock When is “nearest neighbor” meaningful?
\newblock In \emph{Proceedings of the International Conference on Database
  Theory}, pp.\  217--235. Springer, 1999.

\bibitem[Bhatt et~al.(2020)Bhatt, Xiang, Sharma, Weller, Taly, Jia, Ghosh,
  Puri, Moura, and Eckersley]{bhatt_explainable_2019}
Bhatt, U., Xiang, A., Sharma, S., Weller, A., Taly, A., Jia, Y., Ghosh, J.,
  Puri, R., Moura, J.~M., and Eckersley, P.
\newblock Explainable machine learning in deployment.
\newblock In \emph{Proceedings of the 2020 Conference on Fairness,
  Accountability, and Transparency}, pp.\  648--657, 2020.

\bibitem[Doshi-Velez \& Kim(2017)Doshi-Velez and Kim]{doshi-velez_towards_2017}
Doshi-Velez, F. and Kim, B.
\newblock Towards a rigorous science of interpretable machine learning.
\newblock \emph{arXiv preprint arXiv:1702.08608}, 2017.

\bibitem[Dua \& Graff(2021)Dua and Graff]{dua_uci_2020}
Dua, D. and Graff, C.
\newblock {UCI} machine learning repository, 2021.
\newblock URL \url{http://archive.ics.uci.edu/ml}.

\bibitem[Gal \& Ghahramani(2016)Gal and Ghahramani]{gal_dropout_2016}
Gal, Y. and Ghahramani, Z.
\newblock Dropout as a bayesian approximation: Representing model uncertainty
  in deep learning.
\newblock In \emph{Proceedings of the 33th International Conference on Machine
  Learning}, pp.\  1050--1059, 2016.

\bibitem[Goodfellow et~al.(2014)Goodfellow, Shlens, and
  Szegedy]{goodfellow_explaining_2015}
Goodfellow, I.~J., Shlens, J., and Szegedy, C.
\newblock Explaining and harnessing adversarial examples.
\newblock \emph{arXiv preprint arXiv:1412.6572}, 2014.

\bibitem[Guo et~al.(2017)Guo, Pleiss, Sun, and
  Weinberger]{guo_calibration_2017}
Guo, C., Pleiss, G., Sun, Y., and Weinberger, K.~Q.
\newblock On calibration of modern neural networks.
\newblock In \emph{Proceedings of the 34th International Conference on Machine
  Learning}, pp.\  1321--1330, 2017.

\bibitem[Hoffman et~al.(2018)Hoffman, Mueller, Klein, and
  Litman]{hoffman_metrics_nodate}
Hoffman, R.~R., Mueller, S.~T., Klein, G., and Litman, J.
\newblock Metrics for explainable ai: Challenges and prospects.
\newblock \emph{arXiv preprint arXiv:1812.04608}, 2018.

\bibitem[Inselberg(1985)]{inselberg_plane_1985}
Inselberg, A.
\newblock The plane with parallel coordinates.
\newblock \emph{The Visual Computer}, 1\penalty0 (2):\penalty0 69--91, 1985.

\bibitem[Jiang et~al.(2018)Jiang, Kim, Guan, and Gupta]{jiang_trust_2018}
Jiang, H., Kim, B., Guan, M., and Gupta, M.
\newblock To trust or not to trust a classifier.
\newblock In \emph{Advances in Neural Information Processing Systems}, pp.\
  5541--5552, 2018.

\bibitem[Kamar(2016)]{kamar_directions_2016}
Kamar, E.
\newblock Directions in hybrid intelligence: Complementing ai systems with
  human intelligence.
\newblock In \emph{Proceedings of the International Joint Conference on
  Artificial Intelligence}, pp.\  4070--4073, 2016.

\bibitem[Kendall \& Gal(2017)Kendall and Gal]{kendall_what_2017}
Kendall, A. and Gal, Y.
\newblock What uncertainties do we need in bayesian deep learning for computer
  vision?
\newblock In \emph{Advances in Neural Information Processing Systems}, pp.\
  5574--5584, 2017.

\bibitem[Kenny \& Keane(2019)Kenny and Keane]{kenny2019twin}
Kenny, E.~M. and Keane, M.~T.
\newblock Twin-systems to explain artificial neural networks using case-based
  reasoning: comparative tests of feature-weighting methods in ann-cbr twins
  for xai.
\newblock In \emph{Proceedings of the International Joint Conference on
  Artifical Intelligence}, pp.\  2708--2715, 2019.

\bibitem[Lai et~al.(2020)Lai, Carton, and Tan]{lai_harnessing_2020}
Lai, V., Carton, S., and Tan, C.
\newblock Harnessing explanations to bridge ai and humans.
\newblock \emph{arXiv preprint arXiv:2003.07370}, 2020.

\bibitem[Lakkaraju \& Bastani(2020)Lakkaraju and Bastani]{lakkaraju_how_2019}
Lakkaraju, H. and Bastani, O.
\newblock " how do i fool you?" manipulating user trust via misleading black
  box explanations.
\newblock In \emph{Proceedings of the AAAI/ACM Conference on AI, Ethics, and
  Society}, pp.\  79--85, 2020.

\bibitem[Li et~al.(2018)Li, Liu, Chen, and Rudin]{li2018deep}
Li, O., Liu, H., Chen, C., and Rudin, C.
\newblock Deep learning for case-based reasoning through prototypes: A neural
  network that explains its predictions.
\newblock In \emph{Proceedings of the AAAI Conference on Artificial
  Intelligence}, 2018.

\bibitem[Lipton(2018)]{lipton_mythos_2017}
Lipton, Z.~C.
\newblock The mythos of model interpretability.
\newblock \emph{Queue}, 16\penalty0 (3):\penalty0 31--57, 2018.

\bibitem[Lucic et~al.(2020)Lucic, Haned, and de~Rijke]{lucic2020does}
Lucic, A., Haned, H., and de~Rijke, M.
\newblock Why does my model fail? contrastive local explanations for retail
  forecasting.
\newblock In \emph{Proceedings of the 2020 Conference on Fairness,
  Accountability, and Transparency}, pp.\  90--98, 2020.

\bibitem[Lundberg \& Lee(2017)Lundberg and Lee]{lundberg_unified_2017}
Lundberg, S.~M. and Lee, S.-I.
\newblock A unified approach to interpreting model predictions.
\newblock In \emph{Advances in Neural Information Processing Systems}, pp.\
  4765--4774, 2017.

\bibitem[Nushi et~al.(2018)Nushi, Kamar, and Horvitz]{nushi_towards_2018}
Nushi, B., Kamar, E., and Horvitz, E.
\newblock Towards accountable ai: Hybrid human-machine analyses for
  characterizing system failure.
\newblock In \emph{Proceedings of the AAAI Conference on Human Computation and
  Crowdsourcing}, 2018.

\bibitem[Papernot \& McDaniel(2018)Papernot and McDaniel]{papernot_deep_2018}
Papernot, N. and McDaniel, P.
\newblock Deep k-nearest neighbors: Towards confident, interpretable and robust
  deep learning.
\newblock \emph{arXiv preprint arXiv:1803.04765}, 2018.

\bibitem[Rajendran \& LeVine(2019)Rajendran and
  LeVine]{rajendran_accurate_nodate}
Rajendran, V. and LeVine, W.
\newblock Accurate layerwise interpretable competence estimation.
\newblock In \emph{Advances in Neural Information Processing Systems}, pp.\
  13981--13991, 2019.

\bibitem[Ribeiro et~al.(2016)Ribeiro, Singh, and Guestrin]{ribeiro_why_2016}
Ribeiro, M.~T., Singh, S., and Guestrin, C.
\newblock "why should i trust you?" explaining the predictions of any
  classifier.
\newblock In \emph{Proceedings of the 22nd ACM SIGKDD international conference
  on knowledge discovery and data mining}, pp.\  1135--1144, 2016.

\bibitem[Slack et~al.(2020)Slack, Hilgard, Jia, Singh, and
  Lakkaraju]{slack2020fooling}
Slack, D., Hilgard, S., Jia, E., Singh, S., and Lakkaraju, H.
\newblock Fooling lime and shap: Adversarial attacks on post hoc explanation
  methods.
\newblock In \emph{Proceedings of the AAAI/ACM Conference on AI, Ethics, and
  Society}, pp.\  180--186, 2020.

\end{thebibliography}
\bibliographystyle{icml2021}

\end{document}